# Natural Language Understanding with the Quora Question Pairs Dataset

**Lakshay Sharma**
ls4170@nyu.edu

**Laura Graesser**
lhg256@nyu.edu

**Nikita Nangia**
nn1119@nyu.edu

**Utku Evci**
ue225@nyu.edu

## Abstract

This paper explores the task Natural Language Understanding (NLU) by looking at duplicate question detection in the Quora dataset. We conducted extensive exploration of the dataset and used various machine learning models, including linear and tree-based models. Our final finding was that a simple Continuous Bag of Words neural network model had the best performance, outdoing more complicated recurrent and attention based models. We also conducted error analysis and found some subjectivity in the labeling of the dataset.

## 1 Introduction

The Quora dataset is composed of pairs of questions, and the task is to determine if the two questions are duplicates of each other, that is, that they have the same meaning.

Quora is a question answering website where users ask questions and other users respond. The best answers are up-voted and these answers are a valuable learning resource for many topics. Duplicate questions on this site are not uncommon, particularly as the number of questions asked grows. This poses an issue because, if treated independently, duplicate questions may prevent a user from seeing a high quality response that already exists and responders are unlikely to answer the same question twice. Identifying duplicate questions addresses these issues. It reduces the answering burden for responders and makes it possible to direct users to the best responses, improving the overall user experience.

This task requires the model to be successful at Natural Language Understanding (NLU). The task of NLU is to build good representations of human language. It's a challenging and important problem, and success at NLU is necessary to be able to succeed at a host of other Natural Language Processing (NLP) tasks like translation, summarization, and reading comprehension. The problem of determining if two sentences have the same meaning or not requires a model to capture the lexical and syntactic meanings of the sentences presented. Therefore the model must be able to grapple with linguistic phenomena like quantification, tense, modality, and syntactic ambiguity.

Due to the difficulty of this task, we think the Quora dataset poses an interesting problem. In this paper, we aimed to present a comprehensive set of machine learning models, and to study their performance on the dataset. We used simple linear models as our baseline. We built and tested Support Vector Machines (SVM), gradient boosted trees, Random Forests, and a range of deep neural networks. We will discuss all our models in detail and the results we observed.

Duplicate question detection is a binary classification problem on various length strings. The challenging part of the problem is to represent sentences as numerical inputs such that the learning algorithms can work on it. A widely used method involves hand engineered feature generation. This method, combined with tree based models such as random forests, is common in industry. This is the current approach that Quora takes (Dandekar, 2017) and this method can be used together with bag-of-word based models to enhance the performance (Siu, 2016).

With the explosion of neural networks, there has been a lot of work on deep learning methods for sentence classification and building sentence representations (Sutskever et al., 2014; Collobert and Weston, 2008). In particular, there has been work done to succeed at the task of NLU by working on Natural Language Inference (NLI). NLI is a task in which, given a pair of sentences, the model mus determine if the second sentence is entailed, contradicted, or neutral to the first sentence. The Stanford Natural Language Inference corpus (SNLI) brought about a lot of work on the subject (Bowman et al., 2015). Our own neural network explorations took inspiration from for done with SNLI.

## 2 Data

### 2.1 Exploration

The Quora Question Pairs dataset consists of a training set of 404,290 question pairs, and a test set of 2,345,795 question pairs, and is provided as part of a Kaggle competition [1].

Since the test set provided does not contain labels for any question pair, the only measure of performance that can be obtained with this test set is accuracy (via online submission to Kaggle). We therefore felt it better to construct our own test set from the training set provided, since this would allow us to obtain performance metrics other than accuracy, and perform further error analysis of our prediction models. Thus, our data exploration only considered this training set of 404,290 question pairs. More information about how we split this set into training, validation, and test sets is provided in subsection 2.2.

Each sample point has the following fields:

- id: unique ID of each pair
- qid1: ID of first question
- qid2: ID of second question
- question1: text of first question
- question2: text of second question
- is_duplicate: are the questions duplicates of each other (0 indicates not duplicate, 1 indicates duplicate)

Of the 404,290 question pairs, 255,027 (63.08%) have a negative (0) label, and 149,263 (36.92%) have a positive (1) label, making our dataset unbalanced.

While every question pair is unique, every question within the questions pairs is not; 79.22% of questions appear more than once, with one of the questions appearing 158 times. Across all question pairs, there are 537,933 unique questions. Of these, 111,780 questions occur across across multiple pairs. Figure 1 shows the number of times a question appears against the number of questions for that many occurrences.

[1] https://www.kaggle.com/c/quora-question-pairs

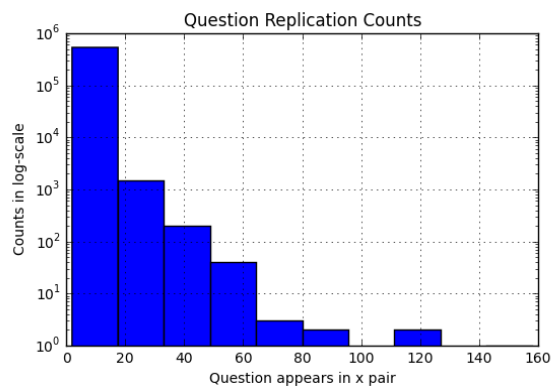

Figure 1: Histogram of question counts

The character set of our dataset was not strictly ASCII; we found that 6,228 questions contained non-ASCII characters, and these questions occurred across 8,744 question pairs. There were also two pairs that contained an empty string for one of their questions.

### 2.2 Splitting

As mentioned in the previous subsection (2.1), while every question pair is unique, a high percentage are a part of multiple pairs.

Therefore, to conduct the data split, we adopted two approaches. For both splits, the provided training data was split into three parts: 70% for training-set, 20% for validation-set, and 10% for test-set.

1. **Blind split**: Split dataset while preserving the balance (the ratio of duplicates to non-duplicates) in each new dataset. For this approach, we ignored the fact that questions in the training set may also appear in the validation and test sets.

2. **Disjoint split**: This split preserves the balance of the data, and it additionally ensures that the questions that appear in the training, validation and test sets are disjoint.

We primarily used the blind split to train and test our models. Unless otherwise specified, all results we present are based on the blind split. The disjoint split makes for a harder task, and may simulate a domain transfer problem. Since the vocabulary of the test and validations sets are different from the vocabulary of the training set, the model needs to learn more generalizable features. Questions tend to be asked more than once, so we feel that the blind data split is more representative of real world applications and therefore our preferred dataset. However we do discuss the performance of our models on the disjoint split in section 4.

### 2.3 Preprocessing

The data is a mass of human generated text and, unsurprisingly, contains anomalies such as non-ASCII characters. Different preprocessing steps we tried eliminate different anomalies, and resultantly lowered the vocabulary size. Without doing anything but tokenizing, we had 175,999 words in our vocabulary. This drops to 94,420 when we remove punctuation and digits.

To understand the effect of different preprocessing steps on the test accuracy, we conducted some experiments. Since the number of experiments grows exponentially with the number of preprocessing functions, we decided to start with experiments with one or a pair of preprocessing functions. We generated pairs like REPLACEPUNC, REMOVE_PUNC or FIXNONASCI, REMOVE_NON_ASCII, such that it logically made sense to pipe them in preprocessing. We ran our experiments three times and calculate the average accuracy on the validation set after training a linear classifier with hinge loss and SGD with $n\_iter = 50$, $\alpha = 0.00005$. We found that even though the vocabulary size falls, the resulting test accuracy is indifferent to different preprocessing steps. Ultimately, we simply used the `nltk.tokenizer` for tokenization as a universal preprocessing step. In addition, for the linear models we removed non-ASCII characters and for the tree-based models we moved punctuation.

## 3 Models

### 3.1 Baseline: Linear Models

Many Statistical Natural Language Processing models make use of n-grams to build up the set of features to be used in a model. Examples of this approach include n-gram language models (Chen and Goodman, 1998), second order Markov Models for part of speech tagging as in (Brants, 2000), and the use of n-character prefixes to build the features for proper name classification. We took the same approach here and built three linear models with different sets of n-gram features.

**Features** Our approach to feature extraction was as follows. Questions were preprocessed by removing the non-ASCII characters, and tokenized. For each data-point, the count of each n-gram in each question was extracted from the preprocessed and tokenized sentences. If the same token appeared in each sentence, this was treated as different feature. This has the advantage that it captures the granular question level n-gram information, which a model can make use of if it is relevant to the problem, and reduces to the word count if the question level information turns out not to be useful.

**Experimental setup** The extracted features were stored in a dictionary representation and finally converted to SciPy's sparse CSR matrix (Jones et al., 2001–) so that it was easily compatible with scikit-learn's implementations of linear and SVM models.

We tested three linear models. In each case the model was the same, logistic regression with L2 regularization, controlled by $\alpha$, trained with stochastic gradient descent using scikit-learn's implementation (Pedregosa et al., 2011). The learning schedule was set to $\eta = 1.0/(\alpha * (t + t_0))$. We tested the following sets of features:

1. Unigram features
2. Unigram and bigram features
3. Unigram, bigram, and trigram features

We roughly tuned the amount of regularization and number of iterations on the unigram model since this was the quickest to train and found that $\alpha = 0.00001$ and 20 iterations gave the best results.

Unsurprisingly, the trigram model gave the best results, since it had the richest feature set, containing all of the features of the unigram and bigram models, plus trigrams. However, it appears to be less efficient than the bigram model, which achieved an accuracy of 79.5% with approximately 1.5m features. Whereas the trigram model, with approximately 2.5m more features only improved over the bigram model by 1.3ppts.

Finally, we tuned our best linear model (the trigram model) and interestingly found that the original settings were almost optimal (see Tables 2 and 3). Reducing the number of iterations to 15 and $\alpha$ to 0.000001 gave marginal improvements to the F1 score on the validation set, but not on the test set.

| Model Class | Model | Test Results Accuracy (%) | F-score |
|---|---|---|---|
| | Most frequent class | 63.1 | - |
| Linear | LR with Unigrams | 75.4 | 63.8 |
| | LR with Bigrams | 79.5 | 70.6 |
| | LR with Trigrams | 80.8 | 71.8 |
| | LR with Trigrams, tuned | 80.1 | 71.5 |
| | SVM with Unigrams | 75.9 | 63.7 |
| | SVM with Bigrams | 79.9 | 70.5 |
| | SVM with Trigrams | 80.9 | 72.1 |
| Tree-Based | Decision Tree | 73.2 | 65.5 |
| | Random Forest | 75.7 | 66.9 |
| | Gradient Boosting | 75.0 | 66.5 |
| Neural Network | CBOW | **83.4** | **77.8** |
| | LSTM | 81.4 | 75.4 |
| | LSTM + Attention | 81.8 | 75.5 |
| | BiLSTM | 82.1 | 76.2 |
| | BiLSTM + Attention | 82.3 | 76.4 |

Table 1: LR in the table stands for logistic regression. The table lists test-set accuracies and F-scores on the blind test-set. The CBOW neural network model attains best accuracy and F-score.

| Regularization weight | Validation Set Results Accuracy (%) | F-score |
|---|---|---|
| 0.1 | 64.0 | 5.7 |
| 0.01 | 71.1 | 45.9 |
| 0.001 | 75.5 | 59.5 |
| 0.0001 | 78.2 | 68.2 |
| 0.00001 | **80.7** | 71.5 |
| 0.000001 | 80.0 | **71.6** |

Table 2: Effect of varying the regularization parameter on the Trigram linear model.

| Number of iterations | Validation Set Results Accuracy (%) | F-score |
|---|---|---|
| 5 | 79.5 | 71.4 |
| 10 | 80.3 | 71.3 |
| 15 | 80.0 | **72.3** |
| 20 | **80.6** | 71.5 |
| 25 | **80.6** | 71.8 |
| 30 | 80.4 | 72.2 |
| 35 | 80.3 | 72.1 |
| 40 | **80.6** | 72.0 |
| 45 | 80.5 | 71.1 |
| 50 | 80.5 | 72.1 |

Table 3: Effect of varying the number of iterations on the Trigram linear model.

### 3.2 Support Vector Machines

Support Vector Machines (SVM) are a popular choice for prediction problems. Even though the optimization done is on a linear loss function, kernelized implementations of SVMs allow separation of data where the separation boundary may be non-linear.

We use scikit-learn's SVM implementation module [2] for our experiments. The features used here are either sparse n-gram matrices, or vector representations of the sentences (sentence embed-dings).

Unless specified otherwise, the following are the default parameters:

- C (penalty parameter C of the error term): 1.0
- kernel (kernel type used in feature matrix): Linear
- gamma (kernel coefficient for RBF, polynomial kernels): $\frac{1}{number\ of\ features}$
- shrinking, whether to use the shrinking heuristic, an optimization to help the models train faster (Bottou and Lin, 2007): true
- max_iter (maximum number of iterations: no limit

---
[2] http://scikit-learn.org/stable/modules/svm.html#svm

| N-gram/kernel | Linear | RBF |
|---|---|---|
| Unigram | 64.2% | 63.1% |
| Bigram | 65.1% | 63.1% |
| Trigram | **65.9%** | - |

Table 4: Accuracy with variation in n-gram features and kernelization (on disjoint validation set)

| N-gram | Accuracy |
|---|---|
| Unigram | 73.5% |
| Bigram | 76.9% |
| Trigram | **78.8%** |

Table 5: Accuracy (linear/no kernelization) with variation in n-gram features (on blind validation set)

**N-gram features** As a first approach, unigram, bigram, and trigram features are used, with linear and RBF kernels, on the disjoint validation dataset; the results are summarized in table 4 (only accuracy was used as a measure of performance here).

Not only does the RBF kernel SVM take a much longer time to train and evaluate, but the accuracy achieved seems lesser than that of the linear kernel SVM, in all cases. For that reason, we only considered the linear kernel for all subsequent experiments.

On both the blind and disjoint datasets, trigram features seemed to outperform unigram and bigram features in terms of accuracy (see tables 4 and 5). This is consistent with our expectations.

**Hyperparameter tuning** We varied the penalty term, $C$, and observed its effect on the accuracy of the model. The max_iter parameter was set to 1000 for this set of experiments. The results for this are summarized in Table 6.

| C value / N-gram | Unigram | Bigram | Trigram |
|---|---|---|---|
| 0.005 | 74.9% | 78.2% | 79.7% |
| 0.01 | 75.3% | 79.0% | **80.4%** |
| 0.1 | 75.5 | 78.9% | 80.1% |
| 0.5 | 74.2% | 77.4% | 79.1% |
| 1.0 | 73.5% | 76.9% | 78.8% |
| 10 | 70.9% | 74.9% | 78.1% |
| 50 | 69.5% | 74.2% | 77.9% |

Table 6: Validation set accuracy with disjoint split with variation in penalty term C, and n-gram features.

Based on further tuning, 0.19 was determined to be the optimal value for penalty term $C$, and trigram features were found to give best performance. The results for the best SVM model on the blind test set are given Table 1, under SVM with trigrams.

**Sentence embedding features** Word embeddings have become a popular and highly effective component of neural network based models in NLP, so we were interested to see if they could be effectively combined with linear models.

For this SVM experiment, 50-dimensional word vectors were obtained using GloVe vectors (6B token version; Pennington et al., 2014). The embedding for a sentence is simply the sum of the embeddings of all the words/tokens that constitute the sentence. These were used these in two ways with the SVM classifier:

- Plain sentence embedding: where the feature for a pair of questions is a 100-dimensional vector, with the first 50 elements representing the embedding of the first question, and the subsequent 50 elements representing the embedding of the second question.

- Distance measures between vectors: for a pair of questions, once the sentence embedding vectors have been retrieved, various distance measures between the two vectors are calculated (using scipy.distance): Bray-Curtis distance, Canberra distance, Chebyshev distance, City Block distance, correlation distance, Cosine distance, Euclidean distance. These seven distance measures are then used as features.

Interestingly, this was the only experiment where the RBF kernel performed better than the linear kernel (results in table 7).

| Feature type/kernel | Linear | RBF |
|---|---|---|
| Sentence embedding vector | 63.0% | **76.9%** |
| Distance measures | 63.6% | 67.5% |

Table 7: Accuracy with variation in penalty term C, and n-gram features (on blind validation set)

### 3.3 Tree-Based Models

Tree based models often give excellent results and are frequently applied in practice. Quora for example currently uses a random forest with hand engineered features on this problem (Dandekar,

2017). For this reason we were interested in applying these models to this problem to understand how they performed and how they differed from linear models and neural networks.

**Feature engineering** It was not possible to use the features from our linear models for our tree based models because it was computationally intractable to train. Models based on decision trees make splitting decisions based on individual features which are sorted to decide a split. This led to two issues. First, to speed up training using the bag of words features, we encoded them in scipy's sparse CSR matrix. However, sparse encodings of features are unsuitable for training tree based models. Splitting decisions are made on individual features but with a sparse encoding, for any split, it is likely that the vast majority of data-points will have no value for the splitting feature and so it will be unclear which partition to assign them too. Theoretically, it is possible to resolve this problem by converting the sparse encoding to a dense matrix. However, for even the unigram model, there were over 175k features. The second issue was that the dense representation took up too much memory and an infeasible amount of to time to train due to the need to sort 175k vectors, of length 280k, to find the optimum split at each node. This forced us to think of alternative ways of generating features.

We examined some examples in the training dataset in order to design a small number of features that could be extracted from every example in the dataset. The final set of features, named Misc, were designed by examining the errors incurred by models trained on the first six sets of features. The full list of initial features is enumerated below and is organized into groups of similar features. To measure the effectiveness of each feature, we tested them on a decision tree, random forest, and a gradient boosted tree, We incrementally added sets of features. We kept the other model parameters constant when tuning for features. Additionally, we removed punctuation from the questions before extracting the features as we found it improved the results.

1. (L) Length based: length for question 1, $\ell_1$, and question 2, $\ell_2$, difference in length, $(\ell_1 - \ell_2)$, and ratio of lengths, $\frac{\ell_1}{\ell_2}$.
2. (LC) Number of common lowercased words: count, count / length pf longest sentence.
3. (LCXS) Number of common lowercased words, excluding stop-words: count, count / length of longest sentence.
4. (LW) Same last word.
5. (CAP) Number of common capitalized words: count, count / length of longest sentence.
6. (PRE) Number of common prefixes, for prefixes of length 3–6: count, count / length of longest sentence.
7. (M) Misc: whether questions 1, 2, and both contain "not", both contain the same digit, and number of common lowercased words after stemming.

Table 4 demonstrates the incremental effectiveness of each of the sets of features. The results are the average of the results from a decision tree, random forest, and gradient boosted tree. Features which counted variants of common words appeared to be the most useful, as did more specific features which identified if the two sentences contain the word "not" or the same digit. Common capitalized words and common prefixes seemed to have less of an impact. However each set of features had some positive effect and so we included them all in the final model.

**Hyperparameters** Finally, we tuned the different parameter settings: maximum depth, minimum number of data-points in a leaf, and the number of estimators for the random forests and gradient boosted trees. This tuning had a negligible impact on the performance of the decision tree model, though it added ∼2% to the accuracy, and increased the F-score by 2–3 points for the random forest and gradient boosted tree models. Our final tree models had the following parameter settings,

1. Decision Tree: Max depth = 10, min samples per leaf = 5
2. Random Forest: Max depth = None, min samples per leaf = 5, num estimators = 50
3. Gradient Boosted Tree: Max depth = 4, n estimators = 500

### 3.4 Neural Networks

There has been a lot of previous work done on paraphrase detection (Socher et al., 2011; Hu et al., 2014), and similar tasks like Natural Language Inference (NLI), in the deep learning world (Rocktäschel et al., 2016; Bowman et al., 2015). The task of detecting identical questions is similar to an NLI task because determining if one question is identical to its pair is like determining if each question is entailed by the other.

| Features | Num features | Acc (%) | Δ acc | F-score | Δ F-score |
|---|---|---|---|---|---|
| Majority classL | 0 | 63.1 | - | - | - |
| L | 4 | 63.7 | 0.6 | 30.7 | - |
| L, LC | 6 | 68.5 | **4.8** | 59.8 | **29.1** |
| L, LC, LCXS | 8 | 70.7 | **2.2** | 63.3 | **3.5** |
| L, LC, LCXS, LW | 9 | 72.7 | 2.0 | 63.6 | 0.3 |
| L, LC, LCXS, LW, CAP | 11 | 72.8 | 0.1 | 64.5 | 0.9 |
| L, LC, LCXS, LW, CAP, PRE | 19 | 73.2 | 0.4 | 64.8 | 0.3 |
| L, LC, LCXS, LW, CAP, PRE, M | 25 | 74.6 | **1.5** | 66.3 | **1.5** |

Table 8: Effects of different hand engineered features on tree based model performance. The results are the average of the results from a decision tree, Random Forest, and gradient boosted machine.

**Model Types** We decided to start with the simplest possible neural network model and gradually try more complicated models. We ultimately tested the following 5 models,

1. Continuous bag of words (CBOW)
2. Long short term memory recurrent neural network (LSTM) (Hochreiter and Schmidhuber, 1997)
3. Bidirection LSTM (BiLSTM)
4. LSTM with word-by-word attention (Bahdanau et al., 2015)
5. BiLSTM with word-by-word attention

All models are built to produce a single vector representation for each question, and use both representations to compute the label prediction. Since we wanted to effectively capture the relationship between the two questions, we concatenated both representations, their difference, and element-wise product (Mou et al., 2016b), and passed this concatenated result to a single `tanh` layer, followed by a softmax classifier. Except in the CBOW model, where the concatenated representation was passed to a deep, 3-layer MLP, followed by a softmax classifier.

For the CBOW model, the vector representation for a question was simple a sum of embedding representations of its words. For all models, to get word embeddings, we used GloVe vectors (840B token version; Pennington et al., 2014). For the LSTM model, we use the final state of the LSTM layer as the vector representation. While for the BiLSTM model, we used an average of all the states of the BiLSTM.

Lastly, for the models with attention, we performed attention by taking the hidden states computed by the LSTM or BiLSTM and inserting into a Parikh et al. (2016) style attention framework. Let's state this mathematically. Let $\{\mathbf{u}_i\}_i^{\ell_u}$ and $\{\mathbf{v}_j\}_j^{\ell_v}$ be the hidden states from the RNN for the two questions, where $\ell_u$ and $\ell_v$ are the lengths of question 1 and 2 respectively. We calculated attention weights, $e_{ij}$, by simply taking a dot product (Luong et al., 2015),

$$e_{ij} = \mathbf{u}_i^T \mathbf{v}_j \quad (1)$$
$$\forall i \in [1, \ldots, \ell_u], \forall j \in [1, \ldots, \ell_v]$$

These weights are normalized and used to get a weighted sum which gives us the new vector representation,

$$\bar{\mathbf{u}}_\mathbf{i} = \sum_{j=1}^{\ell_v} \frac{\exp(e_{ij})}{\sum_{k=1}^{\ell_v} \exp(e_{ik})} \mathbf{v}_j \quad (2)$$

$$\bar{\mathbf{v}}_\mathbf{i} = \sum_{i=1}^{\ell_u} \frac{\exp(e_{ij})}{\sum_{k=1}^{\ell_v} \exp(e_{kj})} \mathbf{u}_i \quad (3)$$

In the LSTM with attention model, the final vector representation was obtained by taking a non-linear combination of the last attention-weighted representation and the last vector representation from the LSTM,

$$\mathbf{u}^* = \tanh(\mathbf{W}^u \bar{\mathbf{u}}_{\ell_u} + \mathbf{V}^u \mathbf{u}_{\ell_u}) \quad (4)$$
$$\mathbf{v}^* = \tanh(\mathbf{W}^v \bar{\mathbf{v}}_{\ell_v} + \mathbf{V}^v \mathbf{v}_{\ell_v}) \quad (5)$$

While in the BiLSTM with attention model, we took a non-linear combination of the average of all attention-weighted representations and the average of all representation from the BiLSTM.

**Experimental setup** As already stated, we used GloVe embeddings to initialize the word embeddings. Out of vocabulary words were initialized with Gaussian samples and all word embeddings were updated during training. The models used

> Labeled as duplicates,
>
> - Q1: How fast are use submarines?, Q2: How fast is a nuclear submarine?
> - Q1: What programming languages do what?, Q2: What do all programming languages do?
>
> Labeled as non-duplicates:
>
> - Q1: Is there any difference between Pepsi and Coca-Cola?, Q2: Whats the difference between Pepsi and Coca-Cola?
> - Q1: What bird is this? Q2: What is this bird?

Figure 2: Examples of pairs of questions with labels that seem incorrect, suggesting subjectivity in the labeling.

300D hidden states. We used Dropout (Srivastava et al., 2014) for regularization on all models, and additional L2 regularization on all but the CBOW model. We used a dropout rate of 0.1, and set the $\beta$ parameter in L2 regularization to 0.01. We used the Adam optimizer (Kingma and Ba, 2015) with default momentum parameters.

## 4 Discussion

It was unsurprising to see that the neural network models outperformed the linear and tree-based models. There are many ways to express a question such that it has the same meaning but uses different words. Neural network models, which make use of word embeddings, are better at modeling similar words, and so we would expect this family of models to perform better than the linear models and tree-based models, which do not represent words in this way.

What was startling to find is that CBOW outstripped the more complicated recurrent models with attention. Not only did CBOW outperform the other models, it edged them out by more than a percentage point. This leads up to believe that for this task the syntactic information in the sentences is far less important than simple word-by-word semantic similarity.

An additional interesting finding is the performance of our models on the disjoint datasets.

Since the disjoint datasets make for a harder task, we expected a drop in performance. The performance of all neural network models dropped by ∼10%. This was to be expected, it's a known problem in NLP that deep models, without additional effort, do not handle domain transfer very well (Mou et al., 2016a).

The linear models experienced an even starker drop in performance. All three types of linear models were barely able to achieve an accuracy above that of the majority class, and the F-scores dropped to 30-40. Since unknown words are ignored by these models, this leads us to think that the vocabulary difference between the training and test set was quite different and indicates that this family of models is incapable of domain transfer.

Specific to the SVM models, it was interesting to observe that RBF kernelization always performed worse than no kernelization (or linear kernelization), except in the cases where features based on sentence embeddings were used. We think that this may be due to the fact that each input vector in our problem has a large number of features. As explained in (Hsu et al., 2003), when the number of features is very large, linear kernels usually work well.

In contrast, the tree based models performed equally well on both data sets. This is to be expected since the hand engineered features are independent of the vocabulary and therefore should be able to handle different vocabulary domains well.

Finally, in doing error analysis, we found that there is some label noise. We discovered examples where the assigned gold label seems incorrect, indicating some subjectivity (see Figure 2 for samples) in the labeling. If this problem is pervasive enough in the data, this may lead to a rather high Bayes error rate.

## 5 Future Work

**Ensemble** The most clear next step for us would be to ensemble the best models of each type (linear, tree-based, and neural network). It's likely that each of this types of models is learning something different about the data and ensembling could definitely increase our accuracy and F-score.

**Feature engineering** Additionally, we could spend more time designing more features for the tree based models. One idea is to use a binary match matrix where the rows represent the first question and the columns represent the second question. We could pick a sentence length size $N$ and initialize the $N \times N$ matrix with zeros. The value $m_{i,j} = 1$ if the words from the two sentences are same $m_{i,j} = -1$ otherwise. If the indices exceed the length of any question, then the entry remains zero. To get more informative features, we could replace the binary counts with normalized relative frequency of words. Even though it is not as powerful as a bag-of-words representation, this approach is somewhat programmatic, making it easier to engineer a larger number of dense features. This represents a compromise between the small number of designed features (25) we used for the tree-based models and the large number of n-gram features which are computationally intractable for trees, and could be used with other dense feature vectors to increase the performance.

**Bayes error rate** To resolve the problem of subjective labeling, it might be worthwhile to conduct a crowd-sourced voting on labels to build better gold labels, and to get an estimate on the human ceiling for the task.

# 6 Conclusion

We tested a large number and variety of machine learning models to solve the duplicate question problem posed by the Quora dataset. Our best performing model was a simple Continuous Bag of Words neural network. We believe the Quora dataset is a useful resource to further explore the task of Natural Language Understanding with machine learning techniques.